\documentclass[lettersize,journal]{IEEEtran}
\usepackage{amsmath,amsfonts}
\usepackage{algorithmic}
\usepackage{algorithm}
\usepackage{array}
\usepackage{textcomp}
\usepackage{stfloats}
\usepackage{url}
\usepackage{verbatim}
\usepackage{graphicx}
\usepackage{subfigure}
\usepackage[compress]{cite} 
\usepackage{float}
\usepackage{booktabs}
\usepackage{multirow}
\usepackage{makecell}
\usepackage{tikz,xcolor}
\usepackage[colorlinks,linkcolor=blue]{hyperref}
\hypersetup{hidelinks,
	colorlinks=true,
	allcolors=black,
	pdfstartview=Fit,
	breaklinks=true}
\definecolor{lime}{HTML}{A6CE39}

\begin{document}
\title{C2F-Thinker: Coarse-to-Fine Reasoning with Hint-Guided Reinforcement Learning for Multimodal Sentiment Analysis}

\author{
	\IEEEauthorblockN{
		Miaosen Luo\IEEEauthorrefmark{1}, 
        Zhenhao Yang\IEEEauthorrefmark{1},
        Jieshen Long\IEEEauthorrefmark{1},
        Jinghu Sun\IEEEauthorrefmark{3},
		Yichu Liu\IEEEauthorrefmark{3},
        Sijie Mai\thanks{† Corresponding author}\IEEEauthorrefmark{1}\IEEEauthorrefmark{2}
        }
        
	\IEEEauthorblockA{\IEEEauthorrefmark{1}School of Computer Science, South China Normal University \\ Guangzhou, Guangdong, China\\}
    \IEEEauthorblockA{\IEEEauthorrefmark{3}School of Electronic and Information Engineering, South China University of Technology \\ Guangzhou, Guangdong, China\\}
    \IEEEauthorblockA{\IEEEauthorrefmark{2}sijiemai@m.scnu.edu.cn}
}

\maketitle
\begin{abstract}
    Multimodal sentiment analysis aims to integrate textual, acoustic, and visual information for deep emotional understanding. Despite the progress of multimodal large language models (MLLMs) via supervised fine-tuning, their "black-box" nature hinders interpretability. While Chain-of-Thought (CoT) reasoning offers a potential remedy, it is constrained by high manual annotation costs and the inherent challenges of reinforcement learning (RL), such as reward sparsity and low exploration efficiency on hard samples. This paper presents C2F-Thinker, a framework that harmonizes coarse-to-fine structured reasoning with hint-guided RL through a two-stage progressive training pipeline. In the first stage, we conduct cold-start supervised fine-tuning using high-quality CoT data distilled from a larger teacher model, consisting of three distinct phases: polarity judgment, intermediate analysis, and fine-grained scoring. This equips the base model with a structured emotional reasoning paradigm. In the second stage, we introduce a hint-guided Group Relative Policy Optimization (GRPO) algorithm. By injecting correct initial polarity predictions as hints during the sampling process, the model is guided toward accurate reasoning paths, effectively mitigating cascading errors and enhancing the utilization of hard samples. Furthermore, a multi-faceted reward function incorporating classification, regression, and formatting constraints is designed to refine prediction accuracy while preserving interpretability. Experimental results demonstrate that C2F-Thinker achieves competitive performance on fine-grained sentiment regression tasks while significantly outperforming baselines in cross-domain generalization. This highlights its potential in building trustworthy and robust sentiment analysis systems for real-world applications.
\end{abstract}

\section{Introduction}
Multimodal Sentiment Analysis (MSA) aims to achieve precise understanding of human emotional states by integrating heterogeneous information from textual, acoustic, and visual modalities \cite{zhang2024comprehensive}. Recently, after extensive supervised fine-tuning (SFT) \cite{luo2025multimodal, gao2025eemo, zhang2025can, chen2024emotionqueen, song2024exploring}, multimodal large language models (MLLMs) have demonstrated superior performance on MSA tasks. However, these models typically operate as "black-box" predictors, where the end-to-end mapping from multimodal inputs to sentiment scores lacks explicit reasoning processes. This opacity greatly limits their application in reliability-sensitive scenarios, such as mental health monitoring and human-computer interaction \cite{khalane2025evaluating}.

To overcome the "black-box" limitation in the decision-making process of pre-trained models, researchers have recently introduced the Chain-of-Thought (CoT) mechanism \cite{wang2025multimodal, zhang2023multimodal, wei2022chain}, aiming to enhance interpretability and logical transparency by explicitly generating intermediate reasoning steps. Nevertheless, traditional CoT reasoning heavily relies on high-quality human-annotated data, which not only incurs high annotation costs but also constrains model performance by the breadth and precision of the annotation distribution. To alleviate this bottleneck, exploring reinforcement learning (RL) for optimizing reasoning paths has become a research hotspot. Among these, Group Relative Policy Optimization (GRPO) \cite{shao2024deepseekmath}, with its advantage of eliminating the need for an additional value network, has become a mainstream paradigm for enhancing complex reasoning capabilities while significantly reducing computational overhead. GRPO constructs reward signals through relative comparisons within a group, partially reducing the dependence on fine-grained external annotations, thus offering a technical path toward building robust interpretable reasoning systems.

\begin{figure}[H]
    \centering
    \includegraphics[width=0.47\textwidth]{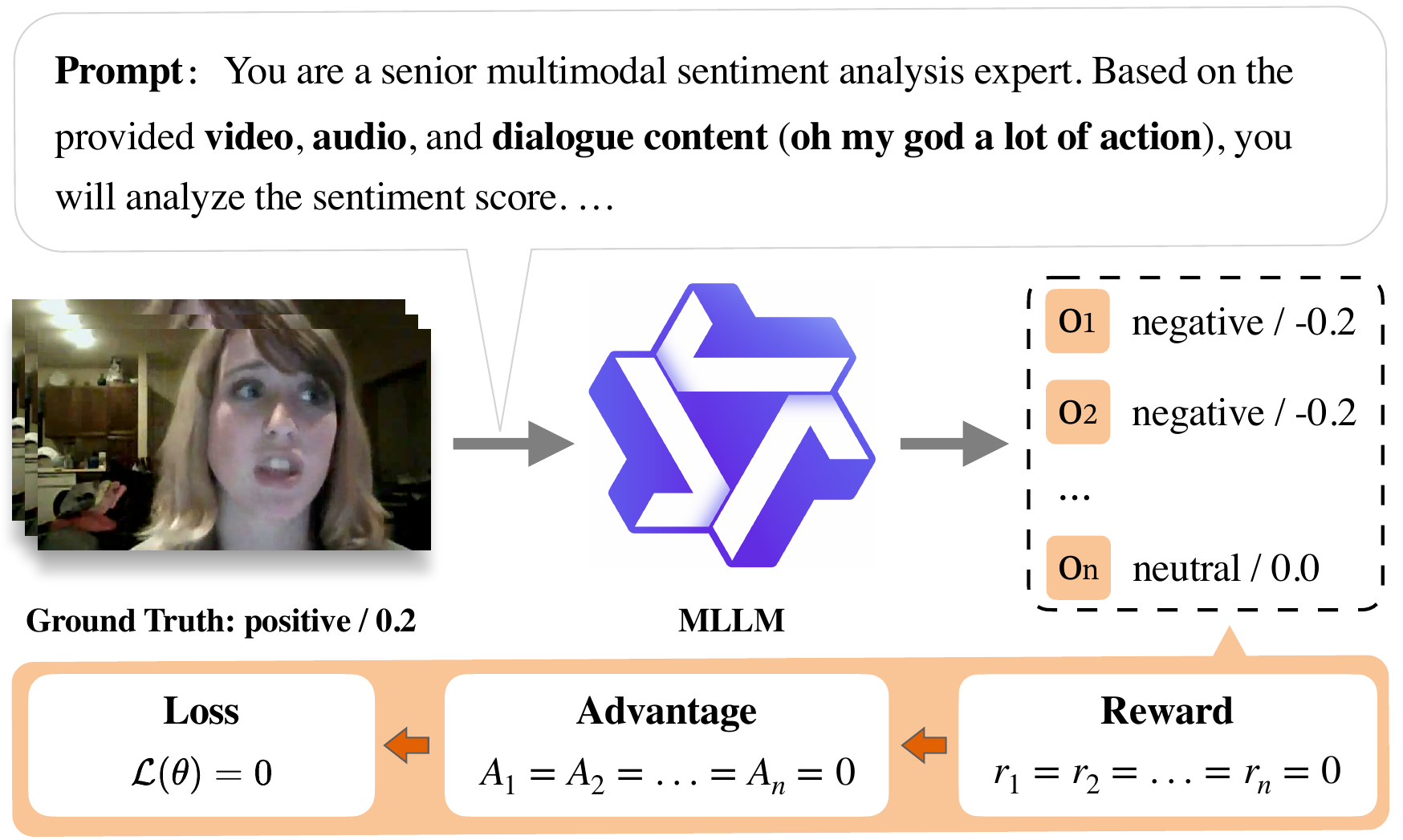}
    \caption{Illustration of the gradient vanishing problem in GRPO when handling hard samples in MSA. }
    \label{fig:hard_sample_example}
\end{figure}

However, directly transferring GRPO to fine-grained sentiment regression tasks faces two fundamental challenges. First, the inherent sparsity of rewards in the continuous regression space poses a challenge for credit assignment. In the traditional chain-of-thought output, the model receives feedback only based on the final sentiment score. Due to the lack of verifiable intermediate nodes, it is difficult to establish robust correlations between the complex reasoning chain and the numerical output, leading to low exploration efficiency in the vast solution space. Second, the "error cascading" effect significantly hinders the utilization of hard samples. In scenarios with subtle emotional cues or modal conflicts, a slight deviation in the initial polarity judgment (coarse-grained) often leads to complete divergence in subsequent logical deduction and numerical estimation. Since the GRPO sampling process lacks guidance mechanisms, the model frequently generates invalid trajectories and fails to extract effective policy gradients from these high-value, challenging samples. As shown in Figure \ref{fig:hard_sample_example}, the MLLM's predictions on hard samples are all incorrect, and the advantages computed by GRPO are all zero, rendering the training samples ineffective.

To address these issues, we propose C2F-Thinker, a novel framework that integrates coarse-to-fine (C2F) structured reasoning with hint-guided GRPO \cite{huang2025boosting}. Our core motivation is to reconstruct the reasoning sequence by introducing verifiable intermediate anchors to alleviate reward sparsity. Specifically, the C2F paradigm requires the model to first perform categorical polarity judgment (e.g., positive, neutral, or negative), then conduct step-by-step analytical reasoning, and finally conclude with a fine-grained sentiment score. Since polarity judgment can be directly verified against the ground-truth label, it serves as a key supervisory signal at an early stage of the output sequence, facilitating more precise reward assignment. This design effectively decomposes the original hard regression problem into a cascade of a classification task and a conditional regression task, thereby reducing the exploration complexity.

Based on this architecture, we implement a two-stage progressive training strategy. In the first stage, we perform C2F-based CoT cold-start SFT. By leveraging a larger teacher model (e.g., Qwen3-Omni-30B \cite{xu2025qwen3}) to generate high-quality reasoning data, we initialize the base model (Qwen2.5-Omni-7B \cite{Qwen2.5-Omni}) to master the structured reasoning format and establish stable baselines for polarity and score prediction. The second stage introduces hint-guided GRPO. For hard samples where the model struggles with initial perception, we provide the ground-truth polarity as a "hint" during the sampling phase to calibrate the reasoning starting point. This guidance effectively steers exploration toward high-reward regions, improving training stability and sample efficiency. The optimization objectives include polarity classification reward, score regression reward, and format consistency reward.

Experimental results on multiple benchmarks show that C2F-Thinker achieves state-of-the-art accuracy in fine-grained regression while maintaining high-quality, structured interpretability. Importantly, our extensive cross-domain evaluation reveals that while direct prediction models suffer a significant performance drop on out-of-distribution (OOD) data, C2F-Thinker exhibits superior generalization by internalizing transferable sentiment logic.

The main contributions of this study are summarized as follows.

\begin{itemize}
    \item We propose the C2F structured reasoning paradigm, which introduces a verifiable polarity anchor to resolve the reward sparsity problem in multimodal regression.
    \item We develop a hint‑guided GRPO algorithm that enhances exploration efficiency on hard samples in MSA tasks by calibrating the initial reasoning trajectory.
    \item We demonstrate through cross-domain experiments that C2F-Thinker exhibits excellent generalization robustness by internalizing transferable sentiment reasoning logic, providing new insights for building trustworthy and transferable multimodal sentiment analysis systems.
\end{itemize}

\section{Related Work}
\subsection{Multimodal Sentiment Analysis}
In recent years, the rise of Multimodal Large Language Models (MLLMs) has led researchers to leverage their powerful comprehension and generation capabilities for sentiment analysis tasks \cite{luo2025multimodal, gao2025eemo, zhang2025can, chen2024emotionqueen, shou2025multimodal}. By performing supervised fine-tuning (SFT) on large-scale multimodal instruction data, MLLMs have demonstrated superior performance in emotion recognition and classification. Nonetheless, these models still operate as "black boxes", taking multimodal data as input and directly outputting sentiment predictions without an interpretable chain of reasoning in between. This opacity limits their practical application in reliability-sensitive scenarios such as mental health monitoring and human-computer interaction. While some studies have attempted to enhance interpretability by introducing emotional reasoning chains, they remain dependent on manually annotated reasoning data, facing challenges such as high labeling costs and restricted reasoning distributions \cite{diwali2023sentiment}.

\begin{figure*}[t]
    \centering
    \includegraphics[width=\textwidth]{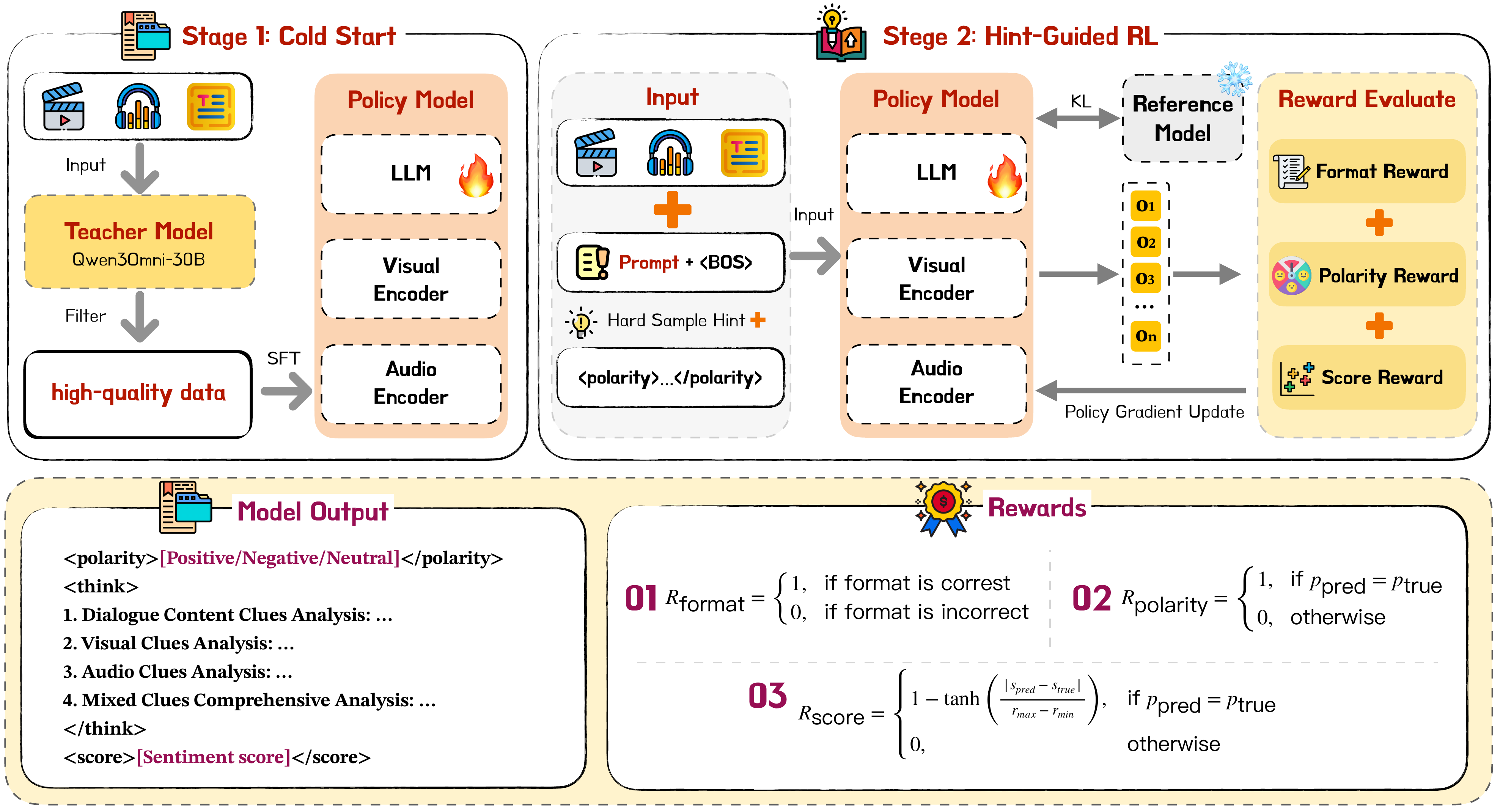}
    \caption{The overall training pipeline for the proposed C2F-Thinker.}
    \label{fig:C2F-Thinker}
\end{figure*}

\subsection{RL for Reasoning Optimization}
To bypass the reliance on expensive manual annotations, recent studies \cite{guo2025deepseek, wen2025reinforcement} have begun exploring the Reinforcement Learning with Verifiable Rewards (RLVR) training paradigm. By utilizing automated reward signals to guide the model's generation of reasoning processes, this technical path provides a new direction for building transparent and interpretable end-to-end MLLMs for multimodal sentiment analysis. Among these, Group Relative Policy Optimization (GRPO) has emerged as a mainstream paradigm for enhancing complex reasoning due to its architectural advantage of eliminating the need for an additional value network. GRPO constructs reward signals by performing relative comparisons among samples within a group, effectively reducing the reliance on fine-grained external annotations. For example, GenCLS++ \cite{he2025gencls++} utilizes RL to constrain the generation space, effectively resolving the challenge of discriminating ambiguous emotional boundaries. CLS-RL \cite{li2025cls} further reveals the synergistic effect of "format rewards + task rewards," proving that models possess the potential to learn classification logic under unsupervised reasoning trajectories. Additionally, AffectGPT-R1 \cite{lian2025affectgpt} validates the effectiveness of this mechanism in open-vocabulary multimodal scenarios, allowing models to spontaneously develop reasoning strategies for complex emotions.

\section{Method}
We propose C2F-Thinker, a two-stage progressive training framework designed to equip MLLMs with both high prediction accuracy and interpretable reasoning capabilities in sentiment analysis tasks. Built upon the Qwen2.5-Omni-7B foundation model, the core innovation of this framework lies in the design of a coarse-to-fine structured reasoning, which guides the model to first discriminate coarse-grained sentiment polarity and then derive fine-grained sentiment scores through calibrated reasoning. This structure is further integrated with the hint-guided GRPO algorithm to synergistically optimize for the challenges of reward sparsity and optimization instability caused by hard samples in fine-grained sentiment prediction tasks. The training pipeline is illustrated in Figure \ref{fig:C2F-Thinker}.

\subsection{Cold Start}

This stage aims to equip the model with a preliminary understanding of structured reasoning for multimodal sentiment analysis.

\subsubsection{Data Generation}
We employ the Qwen3-Omni-30B model as a teacher model to automatically generate CoT reasoning for each sample in the multimodal sentiment dataset. Specifically, by designing prompts, we guide this model to sequentially analyze emotional cues from the textual, audio, and visual modalities, progressively deducing the final sentiment judgment and outputting a regression score. The model adheres to a strict output format, generating sequences containing \texttt{<polarity>}, \texttt{<think>}, and \texttt{<score>} tags. Subsequently, we filter the generated CoT data pairs based on the rationality of the reasoning chain and the consistency between the predicted results and the ground-truth labels. Through this filtering mechanism, we retain high-quality data pairs and replace the generated score within them with the real score label from the dataset. This constructs a training dataset \(\mathcal{D}_{\text{CoT}}\) that possesses both a reliable reasoning process and genuine supervisory signals.

By explicitly including the \texttt{<polarity>} tag within the generated reasoning chain, we obtain a reward signal verifiable early in the generated sequence. This design enables the RL policy to receive immediate feedback at the first step, specifically addressing potential initial polarity prediction errors, especially for hard samples. Based on this, for hard samples, we leverage the ground-truth \texttt{<polarity>} label as hint information during GRPO training to guide the model in generating samples that contribute positively to training during the sampling stage, thereby improving optimization efficiency.

\subsubsection{Training Objective}
The filtered dataset \(\mathcal{D}_{\text{CoT}}\) is used for supervised fine-tuning of the Qwen2.5-Omni-7B base model. The standard next-token prediction loss is applied to the concatenated sequence of the reasoning chain and the final answer, conditioned on the multimodal input \(x\):
\begin{equation}
{
\mathcal{L}_{\text{SFT}} = -\mathbb{E}_{(x, y_{\text{cot}}, y_a) \sim \mathcal{D}_{\text{CoT}}} \left[ \sum_{t} \log p_{\theta}(y_t | x, y_{<t}) \right]
}
\end{equation}
where \(x\) denotes the multimodal input, \(y_{\text{cot}}\) the reasoning chain tokens, \(y_a\) the answer tokens, and \(\theta\) the model parameters. This stage produces a model \(\pi_{\text{SFT}}\) capable of basic interpretable predictions.

\subsection{Hint-guided GRPO Training}

This stage is the core optimization phase of the framework, aiming to systematically enhance the model's reasoning robustness and prediction accuracy through the synergy of GRPO and the Hint mechanism. This stage uses the supervised fine-tuned model \(\pi_{\text{SFT}}\) obtained from the first stage as the initial policy, directly optimizing the generation of sequences containing structured reasoning chains via reinforcement learning.

\subsubsection{Reward Design}
The reward function comprises three components: format reward, polarity reward, and score reward, designed to guide the model to follow the specified output structure, ensure correct macro sentiment, and achieve accurate fine-grained score regression, respectively. The specific design is as follows:
\begin{equation}{
R_{\text{total}} = \lambda_1 R_{\text{format}} + \lambda_2 R_{\text{polarity}} + \lambda_3 R_{\text{score}}
}
\end{equation}
where \(\lambda_1\), \(\lambda_2\) and \(\lambda_3\) are hyperparameters used to balance the contribution of each reward component.

\textbf {Format Reward (\(R_{\text{format}}\))}: Ensures the model strictly adheres to the designed C2F structured reasoning, i.e., generating the complete sequence of \texttt{<polarity>}, \texttt{<think>}, \texttt{<score>} tags in order. Its calculation is as follows:
\begin{equation}{
    R_{\text{format}} = \begin{cases}
    1, & \text{if format is correst } \\
    0, & \text{if format is incorrest}
    \end{cases}
}\end{equation}

\textbf {Polarity Reward (\(R_{\text{polarity}}\))}: This reward encourages the model to make correct macro sentiment polarity judgments. Its calculation is as follows:
\begin{equation}{
    R_{\text{polarity}} = \begin{cases}
    1, & \text{if } p_{\text{pred}} = p_{\text{true}} \\
    0, & \text{otherwise}
    \end{cases}
}\end{equation}
    where \(p_{\text{pred}}\) and \(p_{\text{true}}\) represent the predicted and true sentiment polarity (e.g., ``positive'', ``negative'' or ``neutral'').

\textbf {Score Reward (\(R_{\text{score}}\))}: Under the premise of consistent polarity, it encourages the predicted score to be close to the true value. Its design is as follows:
\begin{equation}{
    R_{\text{score}} = \begin{cases} 1 - \tanh\left(\frac{|s_{\text{pred}} - s_{\text{true}}|}{r_{\text{max}} - r_{\text{min}}}\right), & \text{if } p_{\text{pred}} = p_{\text{true}} \\ 0, & \text{otherwise} \end{cases}
}\end{equation}
    where \(s_{\text{pred}}\) and \(s_{\text{true}}\) represent the predicted and true sentiment scores, and \(r_{\text{max}}\) and \(r_{\text{min}}\) are the upper and lower bounds of score annotations in the dataset.

\subsubsection{Optimization Process}
We employ the GRPO algorithm for RL training of the model. GRPO is an online learning algorithm whose core mechanism is to generate multiple candidate outputs for the same prompt and provide optimization signals through intra-group relative comparisons. Specifically, in each training step, given an input prompt \(x\), the current policy model \(\pi_\theta\) generates a group containing \(G\) candidate sequences, denoted as \(\{o_i\}^{G}_{i=1}\). Subsequently, the total reward \(r_i\) for each sequence \(o_i\) is calculated according to the reward function defined above. To enable ``relative'' comparison and compute optimization signals at each token level, GRPO normalizes the raw rewards within the group to obtain the advantage for each sequence: \(\hat{A}_{i} = \frac{r_i - \text{mean}(\mathbf{r})}{\text{std}(\mathbf{r})}\). The final optimization objective is defined by the following loss function:

\begin{equation}
\resizebox{0.9\linewidth}{!}{%
    $\displaystyle
    \begin{aligned}
    \mathcal{L}_{\text{GRPO}}(\theta) 
    = -\mathbb{E}_{\substack{x \sim \mathcal{D},\\ \{o_i\}_{i=1}^G \sim \pi_{\theta}(\cdot|x)}} 
    &\left[ \frac{1}{G} \sum_{i=1}^G \sum_{t=1}^{|o_i|} \left( \hat{A}_{i,t} \log \pi_{\theta}(o_{i,t} | x, o_{i,<t}) \right. \right. \\
    &\quad \left. \left. - \beta \mathbb{D}_{\text{KL}}[\pi_{\theta}(\cdot | x, o_{i,<t}) \| \pi_{\text{ref}}(\cdot | x, o_{i,<t})] \right) \right]
    \end{aligned}
    $%
}
\end{equation}
where \(\mathbb{D}_{\text{KL}}[\pi_\theta \| \pi_{\text{ref}}]\) is the KL divergence between the current policy and the reference policy \(\pi_{\text{ref}}\), with coefficient \(\beta\) controlling the conservativeness of policy updates.

\begin{table*}[t]
\centering
\caption{Performance comparison on CH-SIMS (OOD) and CH-SIMS v2 (OOD) datasets. All models are trained on the CMU-MOSI dataset. The best results are highlighted in bold, and the runner-up results are indicated with underlines.}
\label{tab:SIMSResult}
    \begin{tabular}{l|c|c|c|c|c|c|c|c|c|c|c|c}
    \hline
    \multirow{2}{*}{Model} & \multicolumn{6}{c|}{CH-SIMS (OOD)} & \multicolumn{6}{c}{CH-SIMSv2 (OOD)} \\ \cline{2-13}
     & Acc5↑ & Acc3↑ & Acc2↑ & F1↑ & MAE↓ & Corr↑ & Acc5↑ & Acc3↑ & Acc2↑ & F1↑ & MAE↓ & Corr↑ \\ 
    \hline
    PandaGPT & 35.2 & 60.6 & 75.1 & 75.1 & 0.488 & 0.488 & 35.6 & 63.9 & 72.9 & 72.4 & 0.453 & 0
    545 \\
    Emotion-LLaMA & 31.1 & 42.0 & 71.8 & 64.1 & 0.481 & 0.485 & 23.7 & 37.0 & 61.7 & 51.7 & 0.475 & 0.464 \\
    MiniCPM-o & 33.7 & 62.6 & 70.5 & 71.6 & 0.481 & 0.575 & \textbf{51.9} & 70.8 & 75.4 & 75.5 & 0.335 & 0.636  \\
    Ola & 38.9 & 69.8 & 77.5 & 78.2 & 0.431 & 0.629 & {48.6} & \underline{76.1} & 80.0 & 80.1 & 0.331 & 0.682 \\
    VideoLLaMA2 & 37.9 & 65.0 & 70.2 & 71.4 & 0.442 & 0.610 & 48.2 & 74.3 & 77.3 & 77.3 & 0.332 & 0.678 \\
    HumanOmni & 38.7 & \underline{70.0} & 77.0 & 77.8 & 0.422 & 0.651 & 45.4 & \textbf {77.2} & \underline{80.4} & \underline{80.5} & 0.342 & 0.707 \\
    Qwen2.5Omni & \underline{47.9} & 67.2 & \underline{81.0} & \underline{78.7} & \textbf {0.352} & \textbf {0.726} & 43.1 & 66.2 & 75.8 & 73.9 & \textbf {0.297} & \underline{0.734} \\
    C2F-Thinker(MOSI) & \textbf {49.0} &\textbf  {73.1} & \textbf {83.8} & \textbf {84.2} & \underline{0.378} & \underline{0.711} & \underline {49.1} & \textbf {77.2} & \textbf {82.7} & \textbf {82.8} & \underline{0.319} & \textbf {0.740} \\
    \hline
    \end{tabular}
\end{table*}

\begin{table*}[t]
\centering
\caption{Performance comparison on CMU-MOSI (ID) and CMU-MOSEI (OOD) datasets. All models are trained on the CMU-MOSI dataset. The best results are highlighted in bold, and the runner-up results are indicated with underlines.}
\label{tab:CMUResult}
    \begin{tabular}{l|c|c|c|c|c|c|c|c|c|c}
    \hline
    \multirow{2}{*}{Model} & \multicolumn{5}{c|}{CMU-MOSI (ID)} & \multicolumn{5}{c}{CMU-MOSEI (OOD)} \\ \cline{2-11}
     & Acc7↑ & Acc2↑ & F1↑ & MAE↓ & Corr↑ & Acc7↑ & Acc2↑ & F1↑ & MAE↓ & Corr↑ \\ 
    \hline
    PandaGPT & 52.1 & 90.2 & 90.2 & \underline{0.536} & \textbf {0.899} & \textbf {50.3} & 85.0 & 85.1 & \textbf {0.587} & \underline{0.758} \\
    Emotion-LLaMA & 40.7 & 86.1 & 86.2 & 0.800 & 0.764 & 41.1 & 83.4 & 83.2 & 2.481 & 0.026 \\
    MiniCPM-o & 49.8 & 89.5 & 89.5 & 0.636 & 0.853 & 33.6 & 86.4 & 86.2 & 0.862 & 0.723 \\
    Ola & 48.3 & 89.3 & 89.3 & 0.620 & 0.860 & 40.4 & 85.2 & 85.4 & 0.670 & 0.743 \\
    VideoLLaMA2 & 50.4 & \underline{90.5} & 90.5 & 0.571 & 0.877 & 44.0 & 84.8 & 84.6 & 0.682 & 0.630 \\
    HumanOmni & \underline{52.8} & \textbf {91.3} & \textbf {91.3} & 0.549 & 0.881 & 46.8 & 86.6 & \underline{86.7} & 0.648 & 0.755 \\
    Qwen2.5Omni & \textbf {53.4} & \underline{90.5} & \underline{90.6} & 0.570 & 0.870 & \underline{47.2} & \underline{86.8} & \underline{86.7} & 0.654 & 0.736 \\
    C2F-Thinker(MOSI) & \underline{52.8} & 88.7 & 88.8 & \textbf {0.521} & \underline{0.893} & 45.7 & \textbf {87.6} & \textbf {87.5} & \underline{0.612} & \textbf {0.780} \\
    \hline
    \end{tabular}
\end{table*}

\subsubsection{C2F-Thinker Implementation}
Our C2F-Thinker aims to solve the problem of sparse rewards when training on hard samples. We provide polarity hints only for hard samples. First, we have the model generate candidate outputs. If the rewards for all outputs fall below a threshold, meaning all are incorrect, the sample is deemed a hard sample. Subsequently, we provide the ground-truth sentiment polarity label (e.g., positive, negative, or neutral) to the model as a hint. This hint guides the model to regenerate the output from the correct starting point. This hint is implemented via in-answer injection, the polarity label is pre-positioned at the beginning of the output sequence, allowing the model to continue writing the reasoning chain and score based on it, thereby ensuring consistency between training and inference input formats.

For hard samples, the hint provides crucial initial correct information, enabling the model to produce partially correct output sequences, thereby obtaining non-zero rewards and effective gradients. This mechanism significantly improves data utilization for hard samples while preventing simple samples from skipping necessary reasoning due to excessive prompting. In this way, the model continuously receives stable optimization signals during reinforcement learning, ultimately improving overall reasoning performance and prediction accuracy.

\section{Experimental Setup}
\subsection{Datasets}
In the task of multimodal sentiment analysis, we select four widely used datasets for training and evaluation. These include CMU-MOSI \cite{zadeh2016multimodal}, CMU-MOSEI \cite{zadeh2018multimodal}, CH-SIMS \cite{yu2020ch}, and CH-SIMSv2 \cite{liu2022make}. To rigorously assess the effectiveness of our proposed reasoning chain method, we examine not only the in-domain performance of the model on each dataset but also its performance under out of distribution (OOD) settings. For example, a model trained on CMU-MOSI is directly evaluated on other datasets. This evaluation design is motivated by a key observation. Without the introduction of a reasoning chain, models tend to overfit the training set. They achieve high performance on the training dataset but exhibit significantly reduced generalization capability on out of domain data with different distributions. By incorporating cross domain evaluation, we can more faithfully reflect the generalization ability and robustness of both the model and the proposed method.

\subsubsection{CMU-MOSI and CMU-MOSEI}
The CMU-MOSI dataset consists of 93 YouTube videos segmented into 2,199 clips. Each clip is annotated with an emotion score on a 7-point scale from strongly negative (-3) to strongly positive (+3). Similarly, the CMU-MOSEI dataset contains 23,453 video clips collected from various online platforms and adopts the same annotation scheme.

\subsubsection{CH-SIMS and CH-SIMSv2}
The CH-SIMS dataset includes 2,281 carefully selected video clips sourced from movies, TV series, and variety shows. Each clip is annotated with an emotion score ranging from negative (-1) to positive (+1). Building on this, the CH-SIMSv2 dataset expands the corpus to include 4,402 supervised clips and 10,161 unsupervised clips (totaling 14,563 clips) from 11 different scenarios, such as vlogs, interviews, and talk shows. While retaining the original annotation methodology, it places greater emphasis on rich non-verbal behaviors.

\subsection{Metrics}
To ensure a fair comparison with traditional research methods, we follow the common practices in the literature for in-domain evaluation by adopting the widely reported standard evaluation metrics for each dataset. Specifically, for the CMU-MOSI and CMU-MOSEI datasets, the evaluation metrics include accuracy for seven discrete sentiment classes (Acc7), accuracy for binary sentiment classification (excluding neutral samples) (Acc2), macro-averaged F1 score (F1), mean absolute error (MAE) between predicted values and ground truth labels, and Pearson correlation coefficient (Corr) between predictions and ground truth. For the CH-SIMS and CH-SIMSv2 datasets, the evaluation metrics include accuracy for five discrete sentiment classes (Acc5), accuracy for three-class classification (positive, neutral, negative) (Acc3), as well as binary classification accuracy (Acc2), F1 score, MAE, and Corr.

\begin{table*}[t]
\centering
\caption{Performance comparison on CH-SIMS (ID) and CH-SIMS v2 (ID) datasets. All models are trained on the CH-SIMS dataset. The best results are highlighted in bold, and the runner-up results are indicated with underlines.}
\label{tab:SIMSResult}
    \begin{tabular}{l|c|c|c|c|c|c|c|c|c|c|c|c}
    \hline
    \multirow{2}{*}{Model} & \multicolumn{6}{c|}{CH-SIMS (ID)} & \multicolumn{6}{c}{CH-SIMSv2 (ID)} \\ \cline{2-13}
     & Acc5↑ & Acc3↑ & Acc2↑ & F1↑ & MAE↓ & Corr↑ & Acc5↑ & Acc3↑ & Acc2↑ & F1↑ & MAE↓ & Corr↑ \\ 
    \hline
    PandaGPT & 38.3 & 58.4 & 77.2 & 74.7 & 0.431 & 0.537 &40&57.6&70.4&67.9&0.403&0.550 \\
    Emotion-LLaMA & 41.1 & 59.3& 77.2 & 75.4 & 0.403 & 0.628 &39.4&57.4&72.3&70.5&0.402&0.610 \\
    MiniCPM-o & 48.8 & 68.7 & 82.5 & 80.5 & 0.350 & 0.695&39.1&68.5&76.6&75.0&0.420&0.700\\
    Ola & 48.4 & 67.8 & 81.6 & 80.2 & 0.406 & 0.646&39.8&70.7&77.6&76.6&0.460&0.645 \\
    VideoLLaMA2 & 52.1 & \underline{74.4} & 81.6 & 82.3 & 0.388& 0.733&\underline{43.3}&\underline{76.4}&80.9&80.9&0.494&0.726\\
    HumanOmni & 52.1 & 72.9& \underline{85.1}& 85.0& \underline{0.327} & 0.749&\underline{43.3}&73.1&81.5&81.3&\underline{0.381}&0.752\\
    Qwen2.5Omni & \textbf {57.6}& 73.3 & \textbf {87.5}& \underline{87.0} & \textbf {0.235} & \textbf {0.841} &\textbf {53.6}&{75.3}&\textbf {85.4}&\textbf {85.2}&\textbf {0.263}&\textbf {0.833}\\
    C2F-Thinker(SIMS) & \underline{54.3} & \textbf{77.2} & \textbf{87.5} & \textbf{87.8} & {0.331} & \underline{0.780} &42.6&\textbf {80.0}&\underline{84.9}&\underline{85.0}&0.407&\underline{0.796}\\
    \hline
    \end{tabular}
\end{table*}

\begin{table*}[t]
\centering
\caption{Performance comparison on CMU-MOSI (OOD) and CMU-MOSEI (OOD) datasets. All models are trained on the CH-SIMS dataset. The best results are highlighted in bold, and the runner-up results are indicated with underlines.}
\label{tab:CMUResult}
    \begin{tabular}{l|c|c|c|c|c|c|c|c|c|c}
    \hline
    \multirow{2}{*}{Model} & \multicolumn{5}{c|}{CMU-MOSI (OOD)} & \multicolumn{5}{c}{CMU-MOSEI (OOD)} \\ \cline{2-11}
     & Acc7↑ & Acc2↑ & F1↑ & MAE↓ & Corr↑ & Acc7↑ & Acc2↑ & F1↑ & MAE↓ & Corr↑ \\ 
    \hline
    PandaGPT & 20.9 & 85.3 & 85.4 & 1.130 & 0.670 &44.6&\underline{85.5}&\underline{85.3}&0.723&0.610 \\
    Emotion-LLaMA & 25.5& 79.2& 79.4& 1.078& 0.593&38.5&80.7&80.5&1.224&0.025\\
    MiniCPM-o & 32.6 & \textbf{87.0} & \textbf{87.1} & 0.874& 0.756&47.1&85.3&84.9&0.641&\underline{0.688}\\
    Ola & 35.2 & 82.7& 82.8 & 0.870 & 0.728 &\underline{47.4}&83.2&83.2&0.666&0.672\\
    VideoLLaMA2 & 25.4 & 69.9 & 69.3 & 1.142 & 0.561&37.7&74.4&72.0&0.835&0.448\\
    HumanOmni & 29.2 & 82.6 & 82.7 & 0.974 & 0.701 &44.0&84.8&84.6&0.682&0.630\\
    Qwen2.5Omni & \underline{39.6} & \underline{85.8} & \underline{85.9} & \underline{0.857} & \underline{0.795} &\textbf {49.3}&84.1&83.4&\underline{0.637}&\underline{0.688}\\
    C2F-Thinker(SIMS) & \textbf {43.2} & {85.0} & {85.1} & \textbf {0.723} & \textbf {0.819} & {46.3} &\textbf {86.6}&\textbf {86.3}&\textbf {0.635} &\textbf {0.731}\\
    \hline
    \end{tabular}
\end{table*}

\subsection{Baselines}
To fairly validate the effectiveness of the reasoning chain method, we select several multimodal large language models capable of simultaneously processing audio, video, and text modalities as baselines for comparison, specifically including PandaGPT \cite{su2023pandagpt}, EmotionLLaMA \cite{cheng2024emotion}, MiniCPM-o \cite{yao2024minicpm}, Ola \cite{liu2025ola}, VideoLLaMA2-AV \cite{zhang2023video}, HumanOmni \cite{zhao2025humanomni}, and Qwen2.5Omni, all with 7B parameters. For these baseline models, we directly employ sentiment scores as labels for supervised fine-tuning, serving as the benchmark for comparison with our proposed method.

\subsection{Implementation Details}
We adopt Qwen2.5-Omni as the base model and first perform supervised fine-tuning using cross-entropy loss with an initial learning rate of \(1 \times 10^{-5}\). In the second-stage reinforcement learning, we set the number of samples to 4, the hyperparameter \(\beta\) to 0.1, and assign equal reward weights of 1. This stage also employs an initial learning rate of \(1 \times 10^{-5}\) with cosine decay scheduling. The entire training process is conducted on 4 NVIDIA A800 GPUs.

\section{Results and Analysis}
\subsection{Main Results}
The performance of C2F-Thinker across four benchmarks is summarized in Tables 1 to 4. In the In-Distribution (ID) scenarios (CH-SIMS and CMU-MOSI), C2F-Thinker consistently achieves state-of-the-art or runner-up performance, proving that the introduction of a coarse-to-fine reasoning chain does not compromise the model's regression precision. Specifically, on the CH-SIMS (ID) dataset, our model achieves the highest Acc3 ($77.2\%$) and F1-score ($87.8\%$), while maintaining a leading position in MAE and Correlation. This indicates that by explicitly modeling the transition from categorical polarity to fine-grained sentiment scores, C2F-Thinker effectively anchors the numerical estimation within a logical emotional framework. Compared to backbone models like Qwen2.5-Omni, which rely on direct label-tuning, our framework demonstrates that structured thinking can match or even exceed the performance of "black-box" predictors in native domain tasks.

The most significant advantage of C2F-Thinker lies in its superior generalization capabilities under OOD conditions. As shown in Tables 2 and 3, when models trained on CH-SIMS are tested on CMU-MOSI/MOSEI (and vice versa), traditional "black-box" MLLMs often suffer from drastic performance degradation due to the domain shift between Chinese and English multimodal data. In contrast, C2F-Thinker exhibits remarkable robustness; for instance, it achieves the best Acc7 ($43.2\%$) and MAE ($0.723$) on CMU-MOSI (OOD) and dominates the CH-SIMS (OOD) benchmarks. This suggests that while direct label-tuning leads to over-fitting on dataset-specific biases, our C2F structured reasoning paradigm encourages the model to internalize transferable sentiment logic. By first identifying a verifiable polarity anchor before proceeding to fine-grained estimation, C2F-Thinker effectively bridges the gap between different data distributions, proving that interpretable reasoning chains are a key driver for cross-domain stability in MSA.

\begin{figure*}[t]
    \centering
    \includegraphics[width=\textwidth]{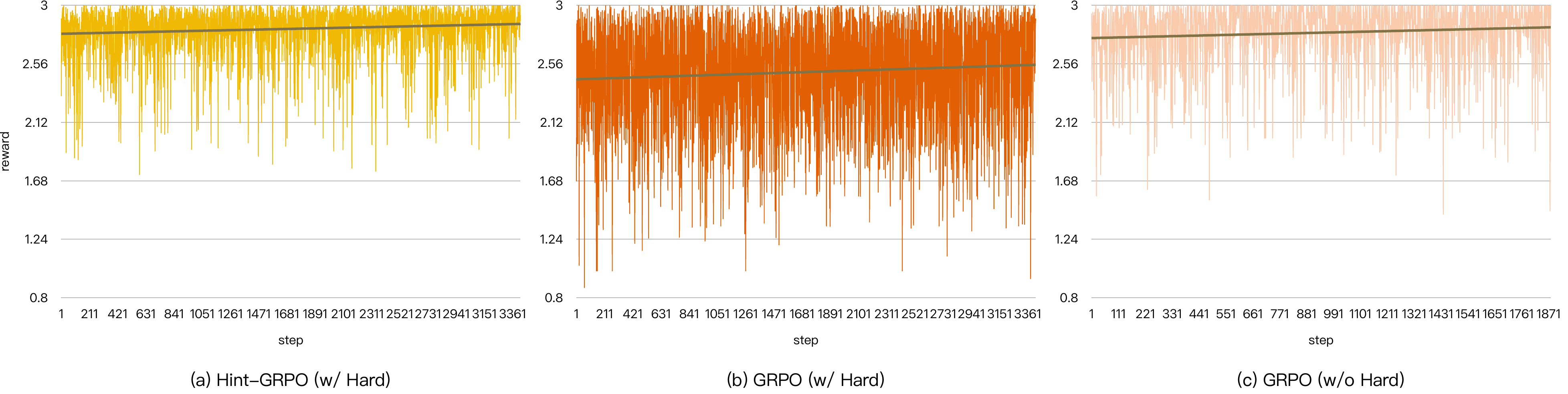}
    \caption{Comparison of reward curves during training for different configurations.}
    \label{fig:rewards}
\end{figure*}

These results fully validate the crucial role of the Chain-of-Thought mechanism in enhancing model generalization. The introduction of a structured reasoning chain compels the model to explicitly analyze and integrate multimodal cues (such as textual semantics, audio prosody, and visual expressions) before generating a final score. This enables the model to learn more universal sentiment representations, rather than simply fitting the statistical distribution of a specific dataset. This reasoning process based on inter-modal correlations allows the model to capture sentiment expression patterns shared across datasets, demonstrating enhanced robustness when facing out-of-domain data with significant distributional differences. It is particularly noteworthy that the slight compromise in regression accuracy observed during in-domain testing with C2F-Thinker is compensated by significant generalization gains in cross-domain scenarios. This phenomenon confirms that the modality interaction logic learned through structured reasoning can be effectively transferred to unknown data distributions, thereby validating the core value of the Chain-of-Thought mechanism in enhancing model generalizability. The comprehensive surpassing of baselines in both classification and regression metrics in cross-domain scenarios proves that the Chain-of-Thought is not a redundant module but a key component for enhancing model robustness and generalization ability.

\subsubsection{Two-Stage Training}
To verify the effectiveness of the two-stage progressive training framework, we compared the performance of the base model under zero-shot prompting, after first-stage SFT, and after second-stage Hint-guided GRPO reinforcement learning optimization. The results are presented in Table \ref{tab:training stages}.

Under the zero-shot setting, the model followed only instruction prompts to generate structured outputs, but all metrics remained low. Despite its general multimodal capabilities, the base model struggled to produce high-quality reasoning chains without fine-tuning, often suffering from format errors or logical inconsistencies, especially under noisy audio or visual inputs.

After first-stage SFT, performance improved substantially. Compared to zero-shot, Acc5 increased by 16.2 percentage points, Acc3 by 21.3 points, Acc2 by 16.7 points and F1 score by 16.1 points. These gains demonstrate that supervised fine-tuning on teacher-generated CoT data enables the model to master structured sentiment reasoning, establish reliable polarity and score predictions, and provide a strong initial policy for reinforcement learning.

After second-stage Hint-guided GRPO optimization, performance improved further. Relative to the SFT version, Acc5 rose by 1.6 points, Acc3 by 1.1 points, Acc2 by 0.8 points, and F1 score by 0.9 points. This confirms that Hint-guided GRPO effectively alleviates reward sparsity in continuous sentiment space by providing polarity hints for hard samples, enabling more efficient policy exploration. Notably, GRPO outperformed SFT across all metrics, indicating that reinforcement learning offers more than repetitive fine-tuning. It enhances discrimination on complex samples, validating the synergy of the two-stage progressive training framework.

\begin{table}[t]
    \caption{Performance comparison of different training stages (Zero-shot, SFT, SFT+RL) on the CH-SIMS dataset.}
    \label{tab:training stages}
    \centering
    \resizebox{0.47\textwidth}{!}{
        \begin{tabular}{c|c|c|c|c|c|c}
        \hline
         & Acc5↑ & Acc3↑ & Acc2↑ & F1↑ & MAE↓ & Corr↑ \\
        \hline
        CoT(Zero-shot) & 36.5 & 55.1 & 70.0 & 70.8 & 0.462 & 0.558 \\
        CoT(SFT) & 52.7 & 76.4 & 86.7 & 86.9 & 0.345 & 0.752 \\
        CoT(SFT+RL) & \textbf {54.3} & \textbf {77.2} & \textbf {87.5} & \textbf {87.8} & \textbf {0.331} & \textbf {0.780} \\
        \hline
        \end{tabular}
    }
\end{table}

\begin{table}[t]
    \caption{Ablation study on the effectiveness of the Hint mechanism on the CH-SIMS dataset.}
    \label{tab:hint}
    \centering
    \resizebox{0.47\textwidth}{!}{
        \begin{tabular}{c|c|c|c|c|c|c}
        \hline
         & Acc5↑ & Acc3↑ & Acc2↑ & F1↑ & MAE↓ & Corr↑ \\
        \hline
        w/o Hint & 46.2 & 67.0 & 82.5 & 81.4 & 0.368 & 0.701 \\
        w/ Hint & \textbf {54.3} & \textbf {77.2} & \textbf {87.5} & \textbf {87.8} & \textbf {0.331} & \textbf {0.780} \\
        \hline
        \end{tabular}
    }
\end{table}

\begin{table}[t]
    \caption{Ablation study on the necessity of hard samples during training on the CH-SIMS dataset.}
    \label{tab:hard sample}
    \centering
    \resizebox{0.47\textwidth}{!}{
        \begin{tabular}{c|c|c|c|c|c|c}
        \hline
         & Acc5↑ & Acc3↑ & Acc2↑ & F1↑ & MAE↓ & Corr↑ \\
        \hline
        w/o Hard & 47.9 & 77.0 & 87.3 & 87.6 & 0.447 & 0.715 \\
        w/ Hard & \textbf {54.3} & \textbf {77.2} & \textbf {87.5} & \textbf {87.8} & \textbf {0.331} & \textbf {0.780} \\
        \hline
        \end{tabular}
    }
\end{table}

\subsection{Ablation Studies}
\begin{figure*}[t]
    \centering
    \includegraphics[width=0.97\textwidth]{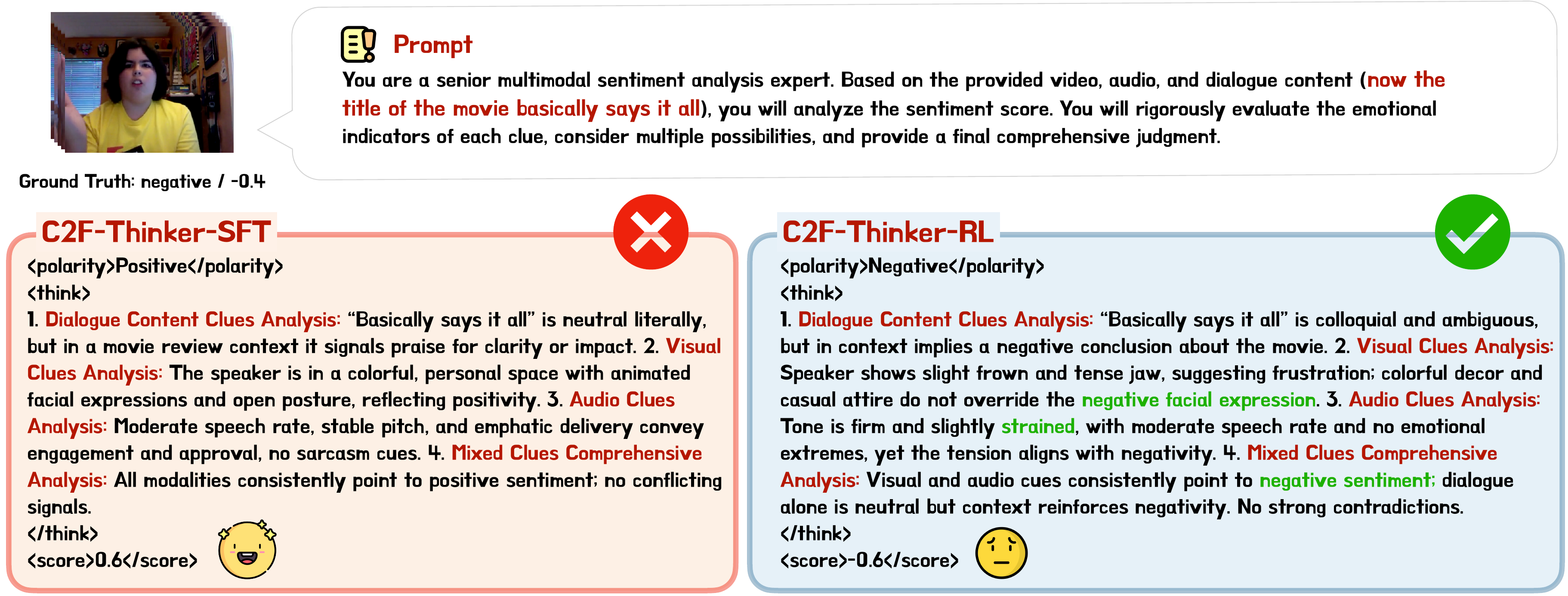}
    \caption{Illustration of C2T-Thinker reasoning.}
    \label{fig:casestudy}
\end{figure*}

\subsubsection{Effectiveness of Hint}
To validate the core role of directional hints in the Hint-guided GRPO mechanism, we conducted an ablation experiment comparing model performance with hints (w/ Hint) and without hints (w/o Hint). Table \ref{tab:hint} reports the results.

It is noteworthy that under the standard GRPO setting without hints, model performance experienced a comprehensive decline compared to the first-stage SFT version. This phenomenon suggests that when the model encounters a large number of challenging samples during training, the extremely low reward signals generated by these samples can lead to chaotic policy gradient directions. This causes the optimization process to fall into local oscillations or even regress, ultimately harming the model's existing performance.

After introducing Hint guidance, model performance was significantly restored and improved. Compared to the w/o Hint version, Acc5 increased by 8.1 percentage points, Acc3 by 10.2 points, Acc2 by 5.0 points, and F1 score by 6.4 points. Moreover, the model with hints comprehensively surpassed the first-stage SFT version, validating the effectiveness of the Hint mechanism. This demonstrates that by providing guiding hints during the polarity discrimination phase for challenging samples, Hint-guided GRPO effectively constrains the exploration range of reinforcement learning within a reasonable region. It transforms potentially harmful noise gradients into beneficial policy updates, thereby enabling continuous optimization of fine-grained sentiment prediction accuracy while maintaining interpretable reasoning.

\subsubsection{Necessity of Incorporating Hard Data}
To investigate the role of challenging samples in reinforcement learning training, we conducted a further ablation experiment comparing two training strategies: GRPO optimization using only easy samples (w/o Hard), and training that included challenging samples with Hint guidance (w/ Hard). The results are shown in Table \ref{tab:hard sample}.

The experimental results indicate that completely excluding challenging samples during the reinforcement learning phase leads to a significant performance gap compared to the version that includes challenging samples supplemented with Hint guidance. Specifically, the w/o Hard version achieved an Acc3 of 77.0\%, slightly lower than the w/ Hard version's 77.2\%; Acc2 and F1 scores were also lower than those of the w/ Hard version; the difference in regression metrics was particularly pronounced: MAE increased from 0.331 to 0.447, and the correlation coefficient decreased from 0.780 to 0.715. This phenomenon suggests that relying solely on easy samples for optimization, while avoiding the risk of reward sparsity from challenging samples, prevents the model from learning the ability to distinguish complex sentiment boundaries from more difficult instances, thereby limiting its generalization performance.

\subsubsection{Training Dynamics Analysis}
To more intuitively demonstrate the optimization process of Hint-guided GRPO, we plotted the reward curves during training in Figure \ref{fig:rewards}. As shown in the figure, in the early stages of training, the reward values for the w/ Hint and w/o Hint versions were similar. As training progressed, the w/o Hint version, lacking directional guidance, exhibited severe reward fluctuations and slow growth when processing challenging samples. In contrast, the reward curve for the w/ Hint version maintained a steady upward trend, converged faster, and ultimately reached a higher stable value. Meanwhile, the w/o Hard version, while avoiding the risk of reward sparsity from challenging samples, showed a relatively stable reward curve with minor fluctuations. However, relying solely on easy samples for optimization prevented the model from learning to distinguish complex sentiment boundaries from more challenging instances, resulting in insufficient exploration and a limited performance ceiling.

Combining this result with the previous ablation experiment (where standard GRPO without hints led to a comprehensive performance decline) yields a more complete conclusion. In reinforcement learning, simply discarding challenging samples is not optimal, and introducing them without guidance can also harm performance. Only by providing effective directional guidance via the Hint mechanism can challenging samples be transformed from an optimization burden into a key performance gain. This enables the model to continuously improve fine-grained sentiment prediction accuracy while maintaining interpretable reasoning, fully validating the necessity and effectiveness of the proposed Hint-guided GRPO mechanism in handling challenging samples.

\subsection{Reasoning Process Analysis}
To illustrate the efficacy of the proposed C2F structured reasoning and hint-guided GRPO, we compare the model outputs before and after RL on a representative hard sample where subtle modal cues conflict with superficial textual content. Following supervised fine-tuning, the model generated a reasoning chain that correctly parsed the literal meaning of the dialogue but over-relied on visual cues such as animated expressions and open posture, leading to a polarity prediction of “Positive” and a score of 0.6 despite the ground-truth label being “Negative” with a score of -0.4. This case exemplifies the “error cascading” issue, where an initial coarse-grained misjudgment, albeit supported by certain modalities, propagates through the reasoning chain to produce a fully divergent numerical output. After reinforcement learning, the same sample elicits a markedly different trajectory the model now identifies the speaker’s slight frown and tense jaw as dominant visual signals, interprets the firm and strained vocal tone as conveying negativity, and recontextualizes the neutral dialogue within this multimodal tension. The polarity is correctly reversed to “Negative”, and the fine-grained score is adjusted to -0.6, which aligns closely with the true sentiment intensity. Critically, the hint-guided GRPO provides the ground-truth polarity as a starting anchor during sampling for such hard samples, enabling the model to bypass initial perceptual biases and explore reasoning paths that consistently integrate conflicting cues. This transition from a superficially coherent but ultimately incorrect chain to a structurally valid and accurate one demonstrates how the verifiable polarity anchor resolves reward sparsity by offering early supervisory signals, while the guided exploration improves sample efficiency on challenging instances, thereby transforming the model from merely mimicking a reasoning format to genuinely internalizing transferable sentiment logic.

\section{Conclusion}
In this paper, we proposed C2F-Thinker, a two-stage progressive training framework that integrates coarse-to-fine structured reasoning with hint-guided reinforcement learning to enhance both interpretability and fine-grained accuracy in MSA. The first stage establishes a structured emotional reasoning paradigm through cold-start supervised fine-tuning using teacher-distilled CoT data, which decomposes sentiment prediction into polarity judgment, intermediate analysis, and fine-grained scoring. The second stage introduces hint-guided GRPO, where ground-truth polarity hints are injected during sampling to calibrate reasoning trajectories, mitigate cascading errors, and improve sample efficiency on hard samples. A multi-faceted reward function further refines prediction accuracy while preserving interpretability. Extensive experiments demonstrate that C2F-Thinker not only achieves competitive performance on fine-grained regression tasks but also significantly outperforms baselines in cross-domain generalization. This work establishes that structured reasoning and hint-guided reinforcement learning can synergistically enhance both model transparency and robustness, offering a systematic pathway toward trustworthy sentiment analysis systems for real-world applications.

\bibliographystyle{IEEEtran}
\bibliography{sample-base}
\vfill
\end{document}